\documentclass[conference]{IEEEtran}
\IEEEoverridecommandlockouts

\usepackage{cite}
\usepackage{amsmath,amssymb,amsfonts}
\usepackage{algorithmic}
\usepackage{graphicx}
\usepackage{textcomp}
\usepackage{xcolor}
\usepackage{multirow}
\usepackage{booktabs}
\usepackage{url}
\DeclareMathOperator*{\RoBERTa}{RoBERTa}

\def\BibTeX{{\rm B\kern-.05em{\sc i\kern-.025em b}\kern-.08em
    T\kern-.1667em\lower.7ex\hbox{E}\kern-.125emX}}
\begin{document}

\title{TraceCoder: Towards Traceable ICD Coding via Multi-Source Knowledge Integration \\
\thanks{This work is supported by Provincial Natural Science Foundation of Jiangsu (NO. BK20250741, NO. BK20240703) , and the Startup Foundation for Introducing Talent of NUIST (NO. 1523142401057, 1523142401055).}
}
\author{
    Mucheng Ren, He Chen, Yucheng Yan, Danqing Hu, Jun Xu, and Xian Zeng$^{\dagger}$~\thanks{Corresponding author: xianzeng@nuist.edu.cn}\\
    \textit{Jiangsu Key Laboratory of Intelligent Medical Image Computing, School of Artificial Intelligence}\\
    Nanjing University of Information Science and Technology, Nanjing, China\\
    \{renm, chenh, 202412491503, danqinghu, jxu, xianzeng\}@nuist.edu.cn
}

\maketitle
\begin{abstract}
Automated International Classification of Diseases (ICD) coding assigns standardized codes to clinical records, playing a critical role in healthcare systems. However, existing methods struggle with semantic gaps between clinical text and ICD codes, poor performance on rare codes, and limited interpretability. We propose TraceCoder, a framework that integrates multi-source external knowledge, including UMLS, Wikipedia, and large language models (LLMs), to enrich ICD code representations and provide traceable, evidence-based predictions. A hybrid attention mechanism is introduced to model interactions among labels, clinical context, and knowledge, improving long-tail code recognition and interpretability by grounding predictions in external evidence. Experiments on MIMIC-III-ICD9, MIMIC-IV-ICD9, and MIMIC-IV-ICD10 datasets demonstrate that TraceCoder achieves state-of-the-art performance, with ablation studies validating its components. TraceCoder offers a scalable, interpretable, and reliable solution for automated ICD coding, aligning with clinical needs for accuracy and traceability.
\end{abstract}


\section{Introduction}
\begin{figure}[t]
    \centering
    \includegraphics[width=\linewidth]{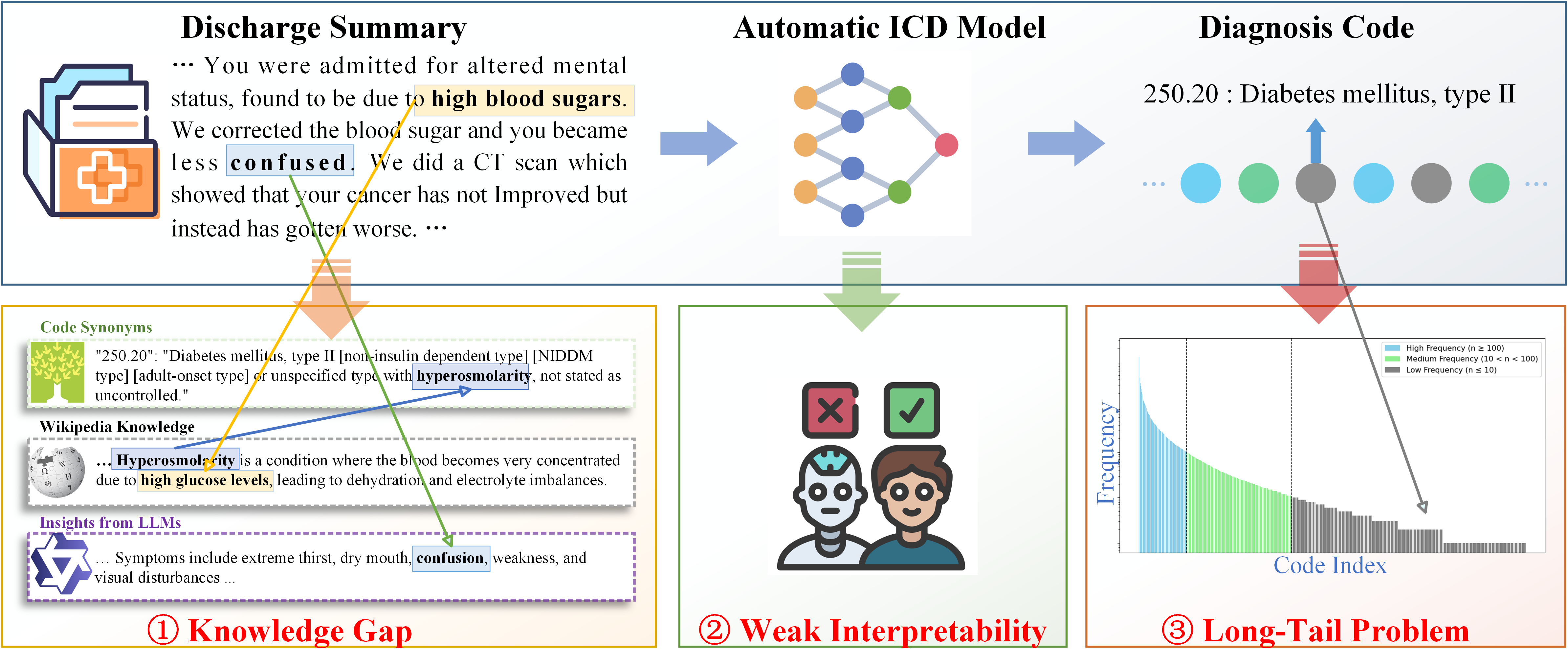}
    \caption{Illustration of Auto-ICD coding task.}
    \label{fig:intro}
\end{figure}

The ICD coding system standardizes diagnoses, supporting billing, epidemiology, and clinical decision-making~\cite{dong2022automated}. Manual ICD coding, however, is labor-intensive, error-prone, and requires extensive expertise. As depicted in Figure~\ref{fig:intro}, automated ICD coding, framed as a multi-label classification task, offers a scalable solution but faces challenges: (1) \textbf{Knowledge Gap}: Clinical notes often lack explicit details, creating semantic gaps between text and corresponding ICD codes. For example, a discharge note may omit the term "Diabetes mellitus" but include lab results such as high blood sugars that imply the condition. (2) \textbf{Weak Interpretability}: The black-box nature of deep learning models makes it difficult for healthcare professionals to trust or verify predictions, especially in critical decision-making. (3) \textbf{Long-tail Problem}: A small number of frequent codes dominate, while most ICD codes are rare. For instance, 54\% of ICD-9 codes in the MIMIC-III-FULL training set are absent from the test set~\cite{edin2023automated}, leading to unreliable predictions for infrequent labels.

\par
Existing studies address these challenges through knowledge enhancement and label attention mechanisms. Knowledge enhancement methods leverage external sources like ICD descriptions, UMLS synonyms, and Wikipedia~\cite{mullenbach-etal-2018-explainable,huang-etal-2022-plm,yuan2022code}, but often fail to integrate complementary information effectively. For example, diabetes (ICD-9 '250.40') and chronic kidney disease (ICD-9 '585.3') lack direct connections in ontologies but are clinically linked~\cite{tuttle2014diabetic}. Meanwhile, LLMs offer precise ICD descriptions, improving code distinguishability, but aligning diverse knowledge sources with clinical text remains challenging.  

\par
Label attention mechanisms learn label-specific vectors by identifying text fragments relevant to specific codes~\cite{vu2021label,yuan2022code,liu2022hierarchical}, improving performance and interpretability. However, these mechanisms lack sufficient traceability, as they fail to link predictions to external evidence—a critical requirement for evidence-based medicine~\cite{sackett1997evidence}.  
\par
To address these limitations, we propose TraceCoder, a traceable automatic ICD coding model via multi-source knowledge integration and attribution. Specifically, instead of simply selecting synonyms as the knowledge to be integrated, a dynamic multi-source knowledge matching module is put forward, which can personalize and incorporate the most relevant knowledge from multi-source knowledge to ensure grounding predictions in external evidence. Additionally, a hybrid attention mechanism is introduced to enhance interactions among label, context and knowledge, which can bridge semantic gaps, mitigate the long-tail problem, and make predictions explainable and traceable through highlighting the clinical text and attributing the external knowledge that influenced the assignment of ICD codes.
\par
We evaluate TraceCoder on MIMIC-III and MIMIC-IV datasets for ICD-9 and ICD-10 coding, achieving state-of-the-art performance. Ablation studies confirm the effectiveness of its components, demonstrating robust performance even in incomplete or ambiguous scenarios.  
\par
Our contributions are as follows:
\begin{enumerate}
\item We propose TraceCoder, a novel framework that integrates dynamic multi-source knowledge for accurate and traceable ICD coding.  
\item A dynamic knowledge matching module aligns external knowledge with clinical context, improving reliability and robustness beyond static synonym-based methods. 
\item A hybrid attention mechanism bridges semantic gaps, mitigates the long-tail problem, and enhances explainability by attributing predictions to key clinical text and external evidence.
\end{enumerate}
\section{Related Works}
Research on automated ICD coding spans attention mechanisms, external knowledge integration, and Transformer-based architectures.
\subsection{Attention Mechanisms for ICD Coding}
Attention mechanisms enhance ICD coding by focusing on relevant text for specific codes. CAML~\cite{mullenbach-etal-2018-explainable} introduced label-wise attention, while LAAT~\cite{vu2021label} modeled text-label dependencies. MSMN~\cite{yuan2022code} and MSAM~\cite{gomes2024accurate} used synonyms to improve code representations but often overlooked external knowledge, limiting performance on rare codes.

\subsection{External Knowledge Integration}
External knowledge bridges semantic gaps in clinical text. Models like MSATT-KG~\cite{xie2019ehr}, BRSK~\cite{zhao2024automated}, and MRR~\cite{wang2024multi} incorporated UMLS, synonyms, and Wikipedia to improve interpretability and handle rare codes. However, they lacked integration between external knowledge and attention mechanisms, failing to capture complex interactions.

\subsection{Transformer-Based Architectures}
Transformer-based models~\cite{vaswani2017attention} capture long-range text dependencies. GatorTron~\cite{yang2022large}, KEPTLongformer~\cite{yang-etal-2022-knowledge-injected}, and PLM-ICD~\cite{huang-etal-2022-plm} adapted pre-trained biomedical models for ICD coding. However, these models are resource-intensive and struggle with the long-tail distribution of ICD codes.

\section{Method}
\begin{figure*}[t]
\centering
    \includegraphics[width=0.9\textwidth]{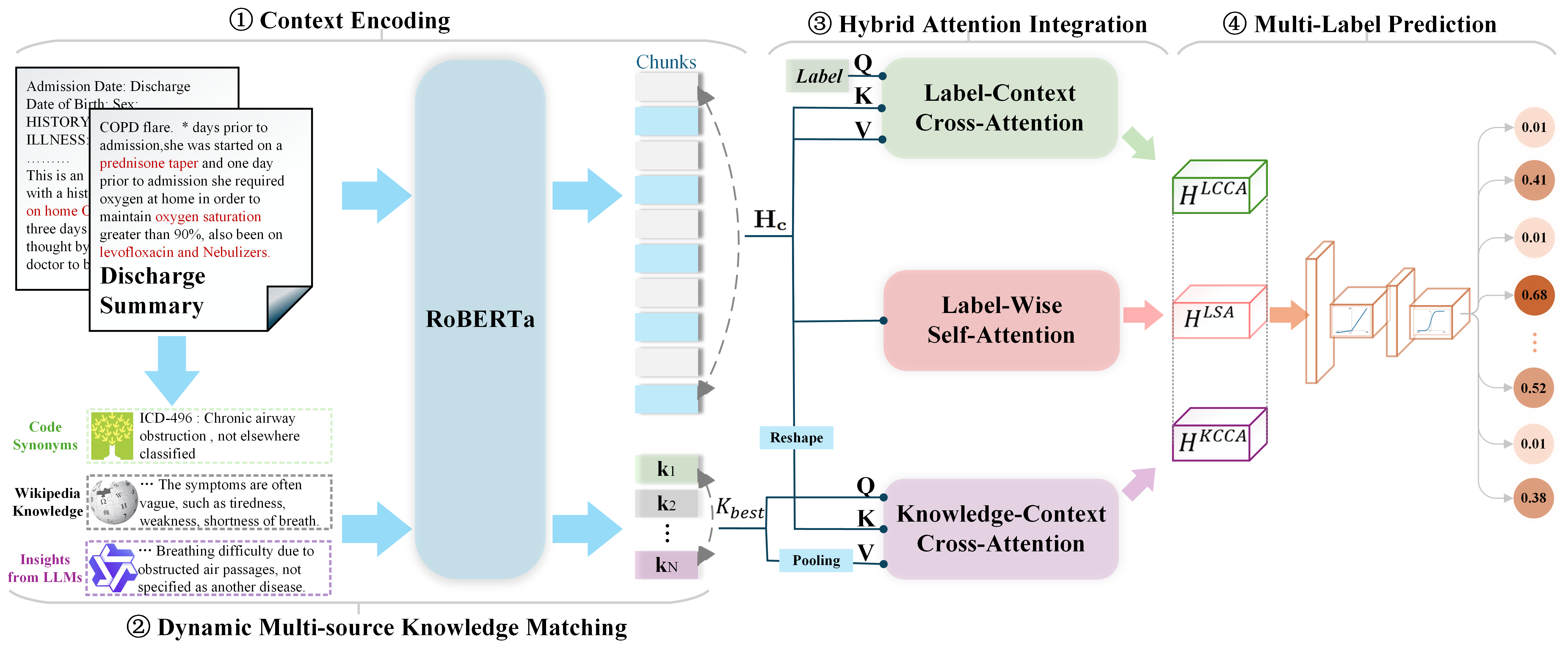}
\caption{The flowchart of proposed TraceCoder framework.} \label{fig:flow}
\end{figure*} 
We frame ICD coding as a multi-label classification task, where the goal is to predict a set of ICD codes from a patient’s discharge summary. The proposed TraceCoder framework addresses this task in four steps, as shown in Figure~\ref{fig:flow}:
\begin{enumerate}
    \item \textbf{Context Encoding}: The input discharge summary is divided into smaller chunks and encoded into contextual representations.
    \item \textbf{Dynamic Multi-source Knowledge Matching}: Relevant clinical knowledge from UMLS, Wikipedia, and LLMs is dynamically aligned and integrated using a pre-trained language model, reducing redundancy and ensuring evidence-grounded predictions.
    \item \textbf{Hybrid Attention Integration}: A hybrid attention mechanism models interactions between clinical text, diagnosis labels, and external knowledge.
    \item \textbf{Multi-Label Prediction}: Final ICD codes are predicted by combining logits from the hybrid attention module.
\end{enumerate}

\subsection{Context Encoding}
We use pre-trained RoBERTa~\cite{lewis-etal-2020-pretrained} as the encoder. Since transformer models are limited to a maximum of $T=512$ tokens, we apply a sliding-window approach to divide the document into $C$ chunks, encoding each chunk individually:

\begin{align}
    \mathbf{H}_c &= \RoBERTa(\mathbf{X}_c), \quad c = 1, 2, \dots, C,
\end{align}

where $\mathbf{X}_c$ represents the token embeddings of the $c$-th chunk, $\mathbf{H}_c$ denotes its contextual representation, and $C$ is the total number of chunks.

\subsection{Dynamic Multi-source Knowledge Matching}
To address the knowledge gaps in clinical text, we propose a dynamic multi-source knowledge matching module that integrates relevant knowledge from external sources to enrich ICD code representations, particularly for rare or ambiguous cases.

\paragraph{Code Synonyms from UMLS}
We enhance ICD code descriptions by extracting synonyms from the UMLS Metathesaurus via Concept Unique Identifiers (CUIs). Synonyms are preprocessed by removing special characters (except hyphens and parentheses) and conjunctions like ``and'' and ``or'' to reduce redundancy. These curated synonyms improve alignment between clinical text and ICD codes by providing medically relevant variations.

\paragraph{Wikipedia Knowledge}
We incorporate additional context from Wikipedia, including synonyms, definitions, symptoms, and disease descriptions. Wikipedia’s detailed and layperson-friendly information helps expand the semantic coverage of ICD codes and ground ambiguous terms in real-world contexts.

\paragraph{Insights from LLMs}
Using large language models (LLMs) like Qwen, we extract disease descriptions, symptoms, and laboratory indicators through tailored prompts. This approach bridges gaps between numerical clinical data (e.g., elevated glucose levels) and corresponding ICD codes (e.g., Type 2 Diabetes), capturing nuanced relationships overlooked by static sources.

To integrate multiple knowledge sources effectively, we represent $N$ candidate knowledge entries (e.g., synonyms or descriptions) as vector embeddings using pre-trained RoBERTa. Each knowledge entry $k_i$ is encoded into a $d$-dimensional vector embedding $\mathbf{k}_i$ from the \texttt{[CLS]} token output:
\begin{align}
    \mathbf{K} = \{\mathbf{k}_1, \mathbf{k}_2, \dots, \mathbf{k}_N\}, \quad \mathbf{k}_i = \RoBERTa(k_i, \texttt{[CLS]}),
\end{align}
where $\mathbf{K}$ is the set of embeddings for the $N$ candidate knowledge entries, and $k_i$ is the $i$-th knowledge entry.

To reduce redundancy and ensure diversity, we solve a \textbf{Maximum Diversity Problem (MDP)}~\cite{glover1977selecting}, selecting a subset $K_{\text{best}} \subseteq \mathbf{K}$ with at most $M$ entries that maximize diversity while preserving semantic coverage. The pairwise dissimilarity between embeddings $\mathbf{k}_i$ and $\mathbf{k}_j$ is measured using cosine distance:
\begin{align}
    d_{ij} = 1 - \frac{\mathbf{k}_i \cdot \mathbf{k}_j}{\|\mathbf{k}_i\| \|\mathbf{k}_j\|},
\end{align}
where $\mathbf{k}_i \cdot \mathbf{k}_j$ is the dot product of the two embeddings, and $\|\mathbf{k}_i\|$ and $\|\mathbf{k}_j\|$ are their Euclidean norms.

The optimization objective is to maximize total pairwise dissimilarity among selected embeddings:

\begin{align}
    \text{maximize} \quad & \sum_{i=1}^{N-1} \sum_{j=i+1}^{N} d_{ij} x_i x_j, \\
    \text{subject to} \quad & \sum_{i=1}^N x_i = M, \\
    & x_i \in \{0, 1\}, \quad \forall i = 1, 2, \dots, N,
\end{align}

where $x_i$ is a binary variable indicating whether the $i$-th embedding is selected ($x_i = 1$). The selected embeddings $K_{\text{best}}$ are given by:
\begin{align}
    K_{\text{best}} = \{\mathbf{k}_i \mid x_i = 1, \, i = 1, 2, \dots, N\}.
\end{align}
The resulting $K_{\text{best}}$ represents the most diverse and relevant subset of knowledge entries, organized as a \textbf{static} matrix for downstream tasks.

\subsection{Hybrid Attention Integration}
The hybrid attention integration component is the core of the TraceCoder framework, combining contextual representations $\mathbf{H}_c$, selected knowledge embeddings $\mathbf{K_{\text{best}}}$, and ICD code embeddings $\mathbf{L}$. It uses three attention mechanisms to model interactions between clinical text, labels, and external knowledge.

\subsubsection{Label-wise Self-Attention (LSA)}
LSA transforms contextual representations $\mathbf{H}_c$ into label-specific vectors by aligning the document with each label in $\mathbf{L}$. Attention weights $\boldsymbol{\alpha}$ are computed as:
\begin{align}
    \boldsymbol{\alpha} = \text{softmax}(\mathbf{W}_2 \tanh(\mathbf{W}_1 \mathbf{H}_c)),
\end{align}
where $\mathbf{W}_1, \mathbf{W}_2$ are learnable matrices. Label-specific representations $\mathbf{H}_c^{\text{LSA}}$ and document-level representations $\mathbf{H}^{\text{LSA}}$ are computed as:
\begin{align}
    \mathbf{H}_{c}^{\text{LSA}} = \mathbf{H}_c \boldsymbol{\alpha}^\top, \quad
    \mathbf{H}^{\text{LSA}} = \sum_{c=1}^C \mathbf{W}_3 \mathbf{H}_c^{\text{LSA}},
\end{align}
where $\mathbf{W}_3$ is a trainable weight matrix and $C$ is the number of chunks.

\subsubsection{Label-Context Cross-Attention (LCCA)}
LCCA models label-document relationships by using the label embeddings $\mathbf{L}$ as queries ($\mathbf{Q}$) and the contextual representation $\mathbf{H}_c$ as keys ($\mathbf{K}$) and values ($\mathbf{V}$):
\begin{align}
    \mathbf{H}_{c}^{\text{LCCA}} = \text{CA}(\mathbf{Q}=\mathbf{L}, \mathbf{K}=\mathbf{H}_c, \mathbf{V}=\mathbf{H}_c),
\end{align}
where $\text{CA}(\cdot)$ is the cross-attention operation. The document-level representation $\mathbf{H}^{\text{LCCA}}$ is obtained by summing over all chunks:
\begin{align}
    \mathbf{H}^{\text{LCCA}} = \sum_{c=1}^C \mathbf{W}_4 \mathbf{H}_c^{\text{LCCA}},
\end{align}
where $\mathbf{W}_4$ is a trainable matrix. The resulting $\mathbf{H}^{\text{LCCA}}$ provides a refined representation of each label, enriched by its interaction with the contextual representation of the clinical document. This alignment strengthens the relationship between labels and clinical text, facilitating accurate label prediction.

\subsubsection{Knowledge-Context Cross-Attention (KCCA)}
The KCCA mechanism integrates external knowledge into the contextual representation of the clinical document. All chunks of the contextual representation $\mathbf{H}_c$ are reshaped into one unified representation $\mathbf{H}$, which serves as the key ($\mathbf{K}$) in the cross-attention operation. The selected knowledge embeddings $\mathbf{K_{\text{best}}}$ act as the query ($\mathbf{Q}$), while their average-pooled embedding $\mathbf{K}_{\text{avg}}$ serves as the value ($\mathbf{V}$).

The cross-attention operation is:
\begin{align}
    \mathbf{H}^{\text{KCCA}} = \text{CA}(\mathbf{Q}=\mathbf{K_{\text{best}}}, \mathbf{K}=\mathbf{H}, \mathbf{V}=\mathbf{K}_{\text{avg}}),
\end{align}
where $\mathbf{H}^{\text{KCCA}}$ is the knowledge-enriched contextual representation. This mechanism aligns clinical context with external knowledge, enhancing semantic understanding for downstream tasks.

\subsection{Multi-Label Prediction}
This component is the final stage of the TraceCoder framework, tasked with predicting a series of target ICD codes from thousands of possible labels. To achieve this, the outputs from the three attention mechanisms, $\mathbf{H}^{\text{LSA}}$, $\mathbf{H}^{\text{LCCA}}$, and $\mathbf{H}^{\text{KCCA}}$, are stacked into a single representation, denoted as $\mathbf{H}_{\text{fused}}$, and processed by a lightweight CNN. The overall computation is expressed as:
\begin{align}
    \mathbf{P} = \sigma(\text{Conv1D}_2(\text{LeakyReLU}(\text{Conv1D}_1(\mathbf{H}_{\text{fused}})))),
\end{align}
where $\mathbf{P}$ represents the final probabilities used for calculating the BCE loss and predicting the target ICD codes.

\begin{table*}[t]
\caption{Comparison of results on the MIMIC-III-50 and MIMIC-III-Full test sets. Bold numbers indicate the best results, and underlined numbers indicate the second best results. * means reproduced by us. $p$-value $<$ 0.01.}
\centering
\resizebox{\textwidth}{!}{%
\begin{tabular}{c|l|ccccc|cccccc}
\hline
\multirow{2}{*}{\textbf{Category}} & \multirow{2}{*}{\textbf{Model}} & \multicolumn{5}{c|}{\textbf{MIMIC-III-50}} & \multicolumn{6}{c}{\textbf{MIMIC-III-FULL}} \\ 
\cline{3-13}
& & \textbf{Micro-AUC} & \textbf{Macro-AUC} & \textbf{Micro-F1} & \textbf{Macro-F1} & \textbf{P@5}  & \textbf{Micro-AUC} & \textbf{Macro-AUC} & \textbf{Micro-F1} & \textbf{Macro-F1} & \textbf{P@8} & \textbf{P@15} \\
\hline
\multirow{2}{*}{Attention-Based Methods} 
& LAAT \cite{vu2021label} & 94.6 & 92.5 & 71.5 & 66.6 & 67.5  & 98.8 & 91.9 & 57.5 & 9.9 & 73.8 & 59.1 \\
& PLM-ICD \cite{huang-etal-2022-plm} & 93.8 & 91.7 & 70.5 & 66.3 & 65.7  & 98.9 & 92.6 & 59.8 & 10.4 & \underline{77.1} & 61.3 \\ 
\hline
\multirow{4}{*}{Knowledge-Based Methods} 
& MSMN \cite{yuan2022code} & 94.4 & 92.5 & 72.5 & 68.3 & 68.0  & \underline{99.2} & 95.0 & 58.4 & 10.3 & 75.2 & 59.9 \\
& FLASH \cite{niu2023retrieve} & - & - & - & - & -  & 99.0 & - & 58.4 & - & 75.8 & 60.2 \\
& BRSK \cite{zhao2024automated} & 94.9 & 92.9 & 72.8 & 69.1 & 67.7  & - & - & - & - & - & - \\
& MRR \cite{wang2024multi} & 94.7 & 92.7 & \underline{73.2} & 68.7 & \underline{68.5}  & \textbf{99.5} & \underline{94.9} & \underline{60.3} & 11.4 & \textbf{77.5} & \underline{62.3} \\ 
\hline
\multirow{4}{*}{Hybrid Methods} 
& KEPTLongformer \cite{yang2022large} & 94.8 & 92.6 & 72.9 & 68.9 & 67.3  & - & - & 59.9 & \underline{11.8} & \underline{77.1} & 61.5 \\
& LCDL \cite{YANG2025103041} & \textbf{95.1} & \underline{93.1} & 73.0 & \underline{69.0} & \textbf{68.6}  & 98.9 & \textbf{95.9} & 59.0 & 11.1 & 75.9 & 60.1 \\
& MSAM(CE)* \cite{gomes2024accurate} & \underline{95.0} & \textbf{93.2} & \underline{73.2} & \underline{69.0} & 68.2  & - & - & - & - & - & - \\
& TraceCoder (Ours) & \textbf{95.1} & \textbf{93.2} & \textbf{73.4} & \textbf{69.1} & \textbf{68.6}  & 98.8 & 92.3 & \textbf{60.5} & \textbf{12.5} & \textbf{77.5} & \textbf{62.4} 
\\ 
\hline
\end{tabular}%
}
\label{tab:results1}
\end{table*}
\begin{table*}[t]
\caption{Comparison of results on the MIMIC-IV-ICD9 and MIMIC-IV-ICD10 test sets. Bold numbers indicate the best results, and underlined numbers indicate the second best results. $^\diamond$ means results are obtained by conducting experiments following the MIMIC-IV data splitting strategy proposed by Nguyen et al.\cite{nguyen2023mimicivicdnewbenchmarkextreme}. $p$-value $<$ 0.01.}
\centering
\resizebox{\textwidth}{!}{%
\begin{tabular}{c|l|cccccc|cccccc}
\hline
\multirow{2}{*}{\textbf{Category}} & \multirow{2}{*}{\textbf{Model}} & \multicolumn{6}{c|}{\textbf{MIMIC-IV-ICD9}} & \multicolumn{6}{c}{\textbf{MIMIC-IV-ICD10}} \\ 
\cline{3-14}
& & \textbf{Micro-AUC} & \textbf{Macro-AUC} & \textbf{Micro-F1} & \textbf{Macro-F1} & \textbf{P@8} & \textbf{P@15} & \textbf{Micro-AUC} & \textbf{Macro-AUC} & \textbf{Micro-F1} & \textbf{Macro-F1} & \textbf{P@8} & \textbf{P@15} \\ 
\hline
\multirow{2}{*}{Attention-Based Methods} 
& LAAT \cite{vu2021label} & 99.3 & 96.0 & 61.7 & 26.4 & 68.9 & 52.7 & 99.0 & 95.4 & 57.9 & 20.3 & 68.9 & 54.3 \\
& PLM-ICD \cite{huang-etal-2022-plm} & 99.4 & \underline{97.2} & \underline{62.6} & 29.8 & 70.0 & 53.5 & \underline{99.2} & 96.6 & \underline{58.5} & \underline{21.1} & 69.9 & \underline{55.0} \\ 
\hline
Knowledge-Based Methods 
& MSMN$^\diamond$ \cite{yuan2022code} & \textbf{99.6} & 96.8 & 61.2 & 27.8 & 68.9 & - & \textbf{99.6} & \textbf{97.1} & 55.9 & 10.8 & 67.7 & - \\ 
\hline
\multirow{2}{*}{Hybrid Methods} 
& CoRelation$^\diamond$ \cite{luo2024corelationboostingautomaticicd} & \underline{99.5} & 96.8 & 62.4 & \underline{30.0} & \underline{70.1} & - & \textbf{99.6} & \textbf{97.2} & 57.8 & 12.6 & \underline{70.0} & - \\
& TraceCoder (Ours) & \underline{99.5} & \textbf{97.5} & \textbf{63.7} & \textbf{33.7} & \textbf{71.1} & \textbf{54.4} & \underline{99.2} & 96.7 & \textbf{60.0} & \textbf{27.7} & \textbf{71.2} & \textbf{56.3} \\ 
\hline
\end{tabular}
}

\label{tab:results2}
\end{table*}

\section{Experiment}

Our experiment was conducted on two publicly available benchmarks MIMIC-III \cite{johnson2016mimic} and MIMIC-IV~\cite{johnson2023mimic} which we accessed through PhysioNet\footnote{\url{https://physionet.org/content/mimiciii/1.4/}} after completing the ethical training program. In this study, we utilize the discharge summaries from the \textit{NOTEEVENTS} table as input text, with the corresponding ICD-9 codes from the \textit{DIAGNOSES-ICD} and \textit{PROCEDURE-ICD} tables serving as the prediction labels. Moreover, for MIMIC-III, we follow Mullenbach et al.\cite{mullenbach-etal-2018-explainable} to form two settings: 1) MIMIC-III-FULL (full codes), 2) MIMIC-III-50 (top 50 frequent codes). For MIMIC-IV, we implement both ICD9 and ICD10 at the same time.

\subsection{Implementation and Evaluation}
We implemented the TraceCoder model on an NVIDIA L40 48G GPU. To prevent overfitting, we applied an early stopping strategy, selecting the model checkpoint with the best performance on the development set. Additionally, the classification threshold was optimized using the development set. Training times per epoch were approximately 8 minutes for MIMIC-III-50, 1 hour for MIMIC-III-FULL, 1.5 hours for MIMIC-IV-ICD9, and 40 minutes for MIMIC-IV-ICD10. For generating code-related insights, we used Qwen-Plus\footnote{\url{https://bailian.console.aliyun.com/model-market/detail/qwen-plus}}. The source code will be made publicly available~\footnote{\url{https://github.com/Hehe517/TraceCoder}}.

For MIMIC-IV experiments, we used the data splits proposed by Edin et al.~\cite{edin2023automated} to ensure consistency with prior work. To fairly compare with baselines, we adopted the evaluation protocol from Edin et al.~\cite{edin2023automated}, which doubles the Macro-F1 scores of published works on MIMIC-III-FULL and MIMIC-IV datasets.

We evaluated the performance of TraceCoder using three common metrics: F1-score (Micro and Macro), AUC (Micro and Macro), and precision at N (P@N).
\subsection{Baselines}
We evaluate TraceCoder against three groups of baseline methods: attention-based methods, knowledge-based methods, and hybrid approaches that integrate both attention mechanisms and external knowledge.

\textbf{Attention-based methods} utilize structured attention mechanisms to capture contextual relationships within clinical text. These approaches emphasize learning representations of clinical notes by focusing on key text fragments to improve ICD code prediction~\cite{vu2021label,huang-etal-2022-plm}.

\textbf{Knowledge-based methods} enrich code representations by integrating external knowledge, without heavily relying on attention mechanisms~\cite{yuan2022code,niu2023retrieve,zhao2024automated,wang2024multi}.

\textbf{Hybrid methods} combine attention mechanisms with external knowledge to address the limitations of single-technique approaches, leveraging both contextual modeling and external information for improved performance~\cite{yang-etal-2022-knowledge-injected,YANG2025103041,gomes2024accurate,luo2024corelationboostingautomaticicd}.
\subsection{Main Results}

Tables~\ref{tab:results1} and~\ref{tab:results2} highlight the superior performance of \textbf{TraceCoder} across the MIMIC-III-50, MIMIC-III-Full, and MIMIC-IV datasets, consistently outperforming state-of-the-art baselines.

On \textbf{MIMIC-III-50}, TraceCoder achieves the highest scores for Micro-F1 (73.4\%), Macro-F1 (69.1\%), Micro-AUC (95.1\%), and Macro-AUC (93.2\%), along with a competitive P@5 (68.6\%), demonstrating its ability to handle both frequent and rare codes effectively.

For \textbf{MIMIC-III-Full}, TraceCoder achieves the best Micro-F1 (60.5\%) and Macro-F1 (12.5\%), with strong precision metrics (P@8: 77.5\%, P@15: 62.4\%), demonstrating its robustness in handling large, imbalanced label distributions.

On \textbf{MIMIC-IV}, TraceCoder excels in both ICD-9 and ICD-10 tasks. It achieves the highest Macro-AUC (97.5\%), Macro-F1 (33.7\%), and Micro-F1 (63.7\%) on ICD-9, and the best Micro-F1 (60.0\%) and Macro-F1 (27.7\%) on ICD-10, alongside consistently strong precision scores.

In summary, TraceCoder outperforms leading baselines by leveraging multi-source knowledge and hybrid attention mechanisms, establishing itself as a robust solution for automated ICD coding across diverse datasets and coding standards.
\section{Discussion}
This section would provide detailed analysis on multi-source knowledge integration and synergistic attention mechanism.

\begin{table}[t]
\centering
\caption{Ablation study results on the MIMIC-III-50 test set.}
\resizebox{\linewidth}{!}{%
\begin{tabular}{lcccccc}

\hline
\multicolumn{1}{l}{\multirow{2}{*}{Ablation Type}} & \multicolumn{1}{c}{\multirow{2}{*}{Method}} & \multicolumn{2}{c}{AUC} & \multicolumn{2}{c}{F1} & \multirow{2}{*}{P@5} \\ \cline{3-6}
\multicolumn{1}{c}{} & \multicolumn{1}{c}{} & Micro & Macro & Micro & Macro &  \\ \hline
\multirow{4}{*}{\begin{tabular}[c]{@{}c@{}}Knowledge\\ Size\end{tabular}}
& M=1 & 94.8 & 93.2 & 72.6 & 69.3 & 67.4 \\
& M=2 & 94.9 & 93.2 & 72.9 & 68.7 & 67.8 \\
& M=4 & 94.8 & 92.8 & 73.0 & 68.3 & 67.6 \\
& M=8 & 95.1 & 93.2 & 73.4 & 69.1 & 68.6 \\ \hline
\multirow{3}{*}{\begin{tabular}[c]{@{}c@{}}Knowledge\\ Source\end{tabular}}
& U & 94.5 & 92.1 & 72.2 & 66.3 & 67.5 \\
& UW & 94.7 & 92.4 & 72.7 & 67.4 & 67.2 \\
& UWQ & 95.1 & 93.2 & 73.4 & 69.1 & 68.6 \\ \hline
\multirow{3}{*}{LLM Source}
& Grok & 94.7 & 92.9 & 72.7 & 69.2 & 67.4 \\
& Deepseek & 94.8 & 93.1 & 73.0 & 69.6 & 67.9 \\
& Qwen & 95.1 & 93.2 & 73.4 & 69.1 & 68.6 \\ \hline
\end{tabular}%
}
\label{tab:abla-combined}
\end{table}

\begin{figure}[t]
\centering
    \includegraphics[width=\linewidth]{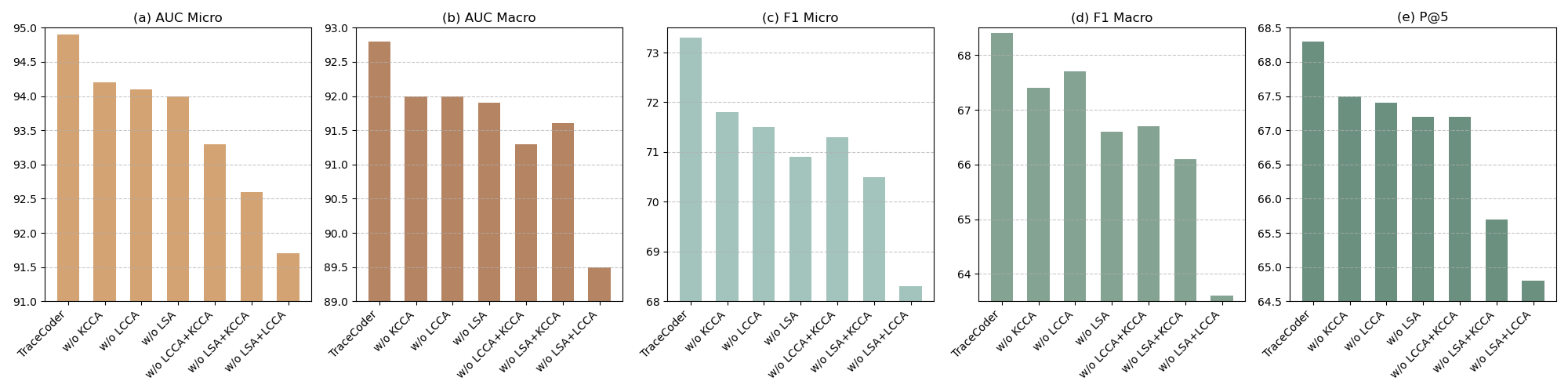}
\caption{Ablation study results on the impact of different attention mechanisms in the MIMIC-III-50 test set.} 
    \label{fig:abla-att}
\end{figure} 

\subsection{Multi-source Knowledge Integration}

Table~\ref{tab:abla-combined} highlights the impact of knowledge size, source, and LLM choice. Increasing knowledge size (\( M \)) improves metrics, with P@5 rising from 67.4\% to 68.6\% and Micro-F1 from 72.6\% to 73.4\%, though Macro-F1 plateaus beyond \( M = 2 \), suggesting diminishing returns and the need for balancing size and efficiency.

Multi-source integration demonstrates complementary benefits. UMLS provides domain-specific knowledge, while Wikipedia adds semantic enrichment, increasing Macro-F1 from 66.3\% (UMLS) to 67.4\% (UMLS + Wikipedia). Adding Qwen LLM further boosts Macro-F1 to 69.1\% and P@5 to 68.6\%, proving the value of combining diverse knowledge sources to bridge semantic gaps and improve rare code predictions.

LLM choice significantly impacts results, with Qwen outperforming Grok and Deepseek, achieving the highest Macro-F1 (69.1\%) and P@5 (68.6\%). Qwen’s precise enrichment enhances TraceCoder’s handling of rare and ambiguous codes, making it the optimal choice.

\subsection{Effect of Hybrid Attention Integration}

Figure~\ref{fig:abla-att} shows the importance of hybrid attention mechanisms. Progressive integration improves metrics, with the full model achieving the best results. LSA captures global label dependencies, LCCA strengthens label-context alignment, and KCCA grounds predictions in external knowledge, bridging semantic gaps and enhancing rare label generalization.

The combination of LCCA and KCCA significantly improves rare label performance, highlighting the power of complementary mechanisms. Together, they enable TraceCoder to achieve state-of-the-art results, improving both frequent and rare label predictions with enhanced robustness and interpretability.

\begin{figure}[t]
\centering
    \includegraphics[width=\linewidth]{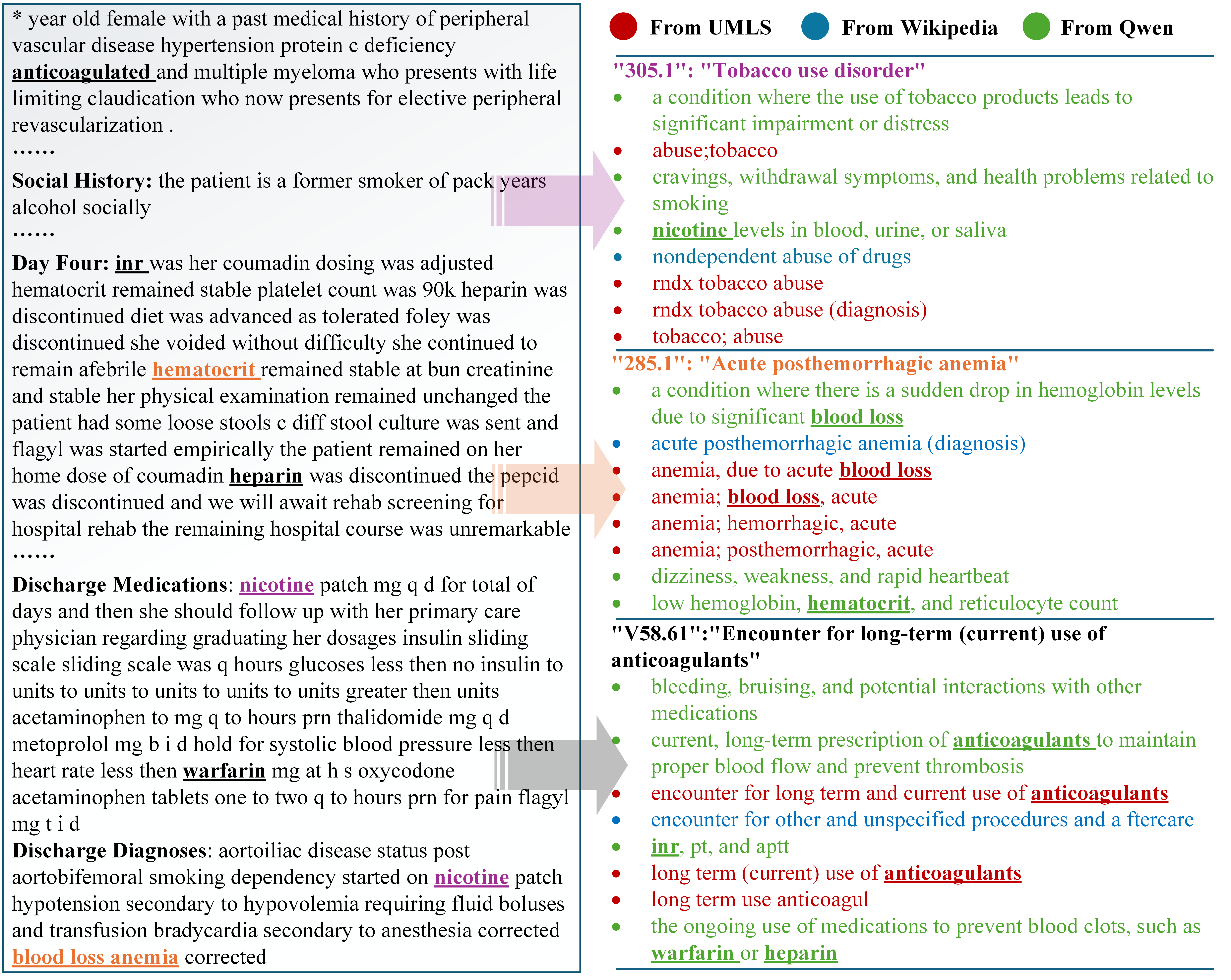}
\caption{Case study illustrating TraceCoder's traceable evidence-grounded prediction.} \label{fig:case}
\end{figure} 
\subsection{Visualization of Traceable Evidence-grounded Prediction}

Figure~\ref{fig:case} illustrates how external knowledge enhances TraceCoder's predictions using a case study from the MIMIC-III dataset. The left side shows a preprocessed clinical note, while the right side presents three target ICD codes and \( M=8 \) knowledge entries, color-coded by source: red (UMLS), blue (Wikipedia), and green (Qwen).

The visualization highlights the complementary contributions of each knowledge source. Overlapping keywords (e.g., ``anticoagulated'' and ``blood loss'') align clinical text with ICD codes, while Qwen provides unique context (e.g., ``nicotine'' and ``hematocrit'') absent in UMLS and Wikipedia. Detailed terms like ``warfarin'' and ``heparin'' demonstrate Qwen’s ability to provide nuanced, context-specific information, improving rare and implicit code predictions.

\section{Conclusion}

We presented \textbf{TraceCoder}, a framework for automated ICD coding that integrates multi-source knowledge with a focus on traceability and explainability. A dynamic knowledge matching module aligns external knowledge (UMLS, Wikipedia, and LLMs) with clinical context, while a hybrid attention mechanism models label-context-knowledge interactions, improving semantic alignment and long-tail code recognition. TraceCoder achieves state-of-the-art results on MIMIC-III and MIMIC-IV datasets. Ablation studies and a case study demonstrate its interpretability and traceability, making it a scalable, robust solution for real-world healthcare systems.

\bibliographystyle{ieeetr}
\bibliography{mybibfile}

\end{document}